\pdfoutput=1
\documentclass[sigconf]{acmart}
\renewcommand\footnotetextcopyrightpermission[1]{} 
\usepackage{booktabs}
\usepackage{tabularx}
\usepackage{etoolbox}

\acmDOI{10.475/123_4}

\acmISBN{123-4567-24-567/08/06}

\acmConference[]{}{}{} 
\acmYear{2017}
\copyrightyear{2017}

\makeatletter
\def\@copyrightspace{\relax}
\makeatother

\patchcmd{\maketitle}{\@copyrightspace}{}{}{}

\begin{document}

\title[DeepSD: Generating High Resolution Climate Change Projections]{DeepSD: Generating High Resolution Climate Change Projections through Single Image Super-Resolution}

\author{Thomas Vandal}
\affiliation{%
  \institution{Northeastern University, Civil and Environmental Engineering}
  \streetaddress{360 Huntington Ave.}
  \city{Boston} 
  \state{MA} 
  \postcode{02141}
}
\email{vandal.t@husky.neu.edu}

\author{Evan Kodra}
\affiliation{%
  \institution{risQ Inc.}
  \streetaddress{404 Broadway}
  \city{Cambridge} 
  \state{MA} 
  \postcode{02139}
}
\email{evan.kodra@risq.io}

\author{Sangram Ganguly}
\affiliation{%
  \institution{Bay Area Environmental Research Institute / NASA Ames Research Center}
  \city{Moffett Field} 
  \state{CA} 
  \postcode{94035}
}
\email{sangram.ganguly@nasa.gov}

\author{Andrew Michaelis}
\affiliation{%
  \institution{University Corporation, Monterey Bay}
  \streetaddress{8 Upper Ragsdale Dr}
  \city{Monterey Bay} 
  \state{CA} 
  \postcode{93940}
}
\email{amac@hyperplane.org}

\author{Ramakrishna Nemani}
\affiliation{%
  \institution{NASA Advanced Supercomputing Division/ NASA Ames Research Center}
  \city{Moffett Field} 
  \state{CA} 
  \postcode{94035}
}
\email{rama.nemani@nasa.gov}

\author{Auroop R Ganguly}
\affiliation{%
  \institution{Northeastern University, Civil and Environmental Engineering}
  \streetaddress{360 Huntington Ave.}
  \city{Boston} 
  \state{MA} 
  \postcode{02141}
}
\email{a.ganguly@neu.edu}

\renewcommand{\shortauthors}{T. Vandal et al.}

\begin{abstract}
The impacts of climate change are felt by most critical systems, such as infrastructure, ecological systems, and power-plants. However, contemporary Earth System Models (ESM) are run at spatial resolutions too coarse for assessing effects this localized. Local scale projections can be obtained using statistical downscaling, a technique which uses historical climate observations to learn a low-resolution to high-resolution mapping. Depending on statistical modeling choices, downscaled projections have been shown to vary significantly terms of accuracy and reliability. The spatio-temporal nature of the climate system motivates the adaptation of super-resolution image processing techniques to statistical downscaling. In our work, we present DeepSD, a generalized stacked super resolution convolutional neural network (SRCNN) framework for statistical downscaling of climate variables. DeepSD augments SRCNN with multi-scale input channels to maximize predictability in statistical downscaling. We provide a comparison with Bias Correction Spatial Disaggregation as well as three Automated-Statistical Downscaling approaches in downscaling daily precipitation from 1 degree (~100km) to 1/8 degrees (~12.5km) over the Continental United States. Furthermore, a framework using the NASA Earth Exchange (NEX) platform is discussed for downscaling more than 20 ESM models with multiple emission scenarios.
\end{abstract}



\maketitle


\section{Introduction}
Climate change is causing detrimental effects to society's well being as temperatures increase, extreme events become more intense\cite{pachauri2014climate}, and sea levels rise\cite{nicholls2010sea}. Natural resources that society depends on, such as agriculture, freshwater, and coastal systems, are vulnerable to increasing temperatures and more extreme weather events. Similarly transportation systems, energy systems, and urban infrastructure allowing society to function efficiently continue to degrade due to the changing climate. Furthermore, the health and security of human beings, particularly those living in poverty, are vulnerable to extreme weather events with increasing intensity, duration, and frequency~\cite{trenberth2012framing}. Scientists and stakeholders across areas such as ecology, water, and infrastructures, require access to credible and relevant climate data for risk assessment and adaptation planning.

Earth System Models (ESMs) are physics-based numerical models which run on massive supercomputers to project the Earth’s response to changes in atmospheric greenhouse gas (GHG) emissions scenarios. Archived ESM outputs are some of the principal data products used across many disciplines to characterize the likely impacts and uncertainties of climate change~\cite{taylor2012overview}. These models encode physics into dynamical systems coupling atmospheric, land, and ocean effects. ESMs provide a large number of climate variables, such as temperature, precipitation, wind, humidity, and pressure, for scientists to study and evaluate impacts. The computationally demanding nature of ESMs limits spatial resolution between 1 and 3 degrees. These resolutions are too course to resolve critical physical processes, such as convection which generates heavy rainfall, or to assess the stakeholder-relevant local impacts of significant changes in the attributes of these processes~\cite{schiermeier2010real}.

Downscaling techniques are used to mitigate the low spatial resolution of ESMs through dynamical and statistical modeling. Dynamical downscaling, also referred to as regional climate models (RCMs), account for local physical processes, such as convective and vegetation schemes, with sub-grid parameters within ESM boundary conditions for high-resolution projections. Like ESMs, RCMs are computationally demanding and are not transferable across regions. In contrast, the statistical downscaling (SD) technique learns a functional form to map ESMs to high resolution projections by incorporating observational data. A variety of statistical and machine learning models, including linear models~\cite{hessami2008automated}, neural networks~\cite{cannon2011quantile}, and support vector machines~\cite{Ghosh2010}, have been applied to SD and is discussed further in section 4 (Related Work). Despite the availability of many techniques, we are not aware of any SD method which explicitly captures spatial dependencies in both low-resolution and high-resolution climate data. Furthermore, traditional methods require observational data at the high-resolution target, meaning that regions with little observation data, often the poorest regions which are most effected by climate change, are unable to receive downscaled climate data needed for adaptation. 

The lack of explicit spatial models in SD of ESMs motivated us to study the applicability of computer vision approaches, most often applied to images, to this problem. More specifically, advances in single image super-resolution (SR) correspond well to SD, which learn a mapping between low- and high-resolution images. Moreover, as SR methods attempt to generalize across images, we aim to provide downscaled climate projections to areas without high-resolution observations through what may be thought of as transfer learning. Though we will discuss this topic further in section 4 (Related Work), we found that super-resolution convolutional neural networks were able to capture spatial information in climate data to improve beyond existing methods.

Lastly, we present a framework using our super-resolution approach to downscale ensemble ESMs over the Continental United States (CONUS) at a daily temporal scale for four emission scenarios by using NASA's Earth Exchange (NEX) platform. NEX provides a platform for scientific collaboration, knowledge sharing and research for the Earth science community. As part of NEX, along with many other earth science data products, NASA scientists have already made monthly downscaled ESMs for CONUS up to the year 2100 at 30 arc seconds (NEX-DCP30) that are openly available to the public. However, the downscaling methodology, bias correction spatial disaggregation, has limitations and the monthly scale reduces the applicability to studying extreme events. The improvement of such data products is vital for scientists to study local impacts of climate change to resources society depends on. 

\subsection{Key Contributions}
The key contributions are as follows:
\vspace{-1mm}
\begin{itemize}
  \item We present DeepSD, an augmented stacked super-resolution convolutional neural network for statistical downscaling of climate and earth system model simulations based on observational and topographical data.
  \item DeepSD outperforms a state-of-the-art statistical downscaling method used by the climate and earth science communities as well as a suite of off-the-shelf data mining and machine learning methods, in terms of both predictive performance and scalability. 
  \item The ability of DeepSD to outperform and generalize beyond grid-by-grid predictions suggests the ability to leverage cross-grid information content in terms of similarity of learning patterns in space, while the ability to model extremes points to the possibility of improved ability beyond matching of quantiles. Taken together, this leads to the new hypothesis that methods may be able to use spatial neighborhood information to predict in regions where data may be sparse or low in quality.
  \item For the first time, DeepSD presents an ability to generate, in a scalable manner, downscaled products from model ensembles, specifically, simulations from different climate modeling groups across the world run with different emissions trajectories and initial conditions.
  \item DeepSD provides NASA Earth Exchange (NEX) a method of choice to process massive climate and earth system model ensembles to generate downscaled products at high resolutions which can then be disseminated to researchers and stakeholders. 
\end{itemize}

\subsection{Organization of the Paper}

The remainder of the paper is organized as follows. Section 2 (Earth Science Data) presents necessary data used for SD along with their associated challenges. Section 3 (Statistical Downscaling) discusses the problem of SD. Section 4 (Related Work) discusses techniques previously applied to SD along with an overview of super-resolution methods. Furthermore, we discuss the relationships between images and climate data. Section 5 (Methodology) presents DeepSD, the augmented stacked super-resolution convolutional neural network formulation. In section 6 (Experiments) we compare our method to another SD technique and three off-the-shelf machine learning approaches and outline the process by which we will scale our method to many climate model simulations. In section 7 (conclusion) we briefly discuss results, limitations, and future work.

\section{Earth Science Data}

Earth science data stems from a variety of areas, including climate simulations, remote sensing through satellite observations, and station observations. The spatio-temporal nature of such data causes heavy computational and storage challenges. For instance, a single climate variable at the daily temporal and 4km spatial scales over only the United States requires 1.2GB of storage. Multiplying this effect over a large number of variables, including precipitation, temperature, and wind, globally creates high storage and processing requirements. Furthermore, analysis of these complex datasets require both technical and domain expertise.

ESM outputs, as discussed previously, are one form of earth science data which is crucial to the understanding of our changing climate. The most recent ESMs are a product of the fifth phase of the Coupled Model Intercomparision Project which simulate the climate through a dynamical system coupling effects from the atmosphere, land, and ocean~\cite{taylor2012overview}. However, it is well understood that holes in these models exist, including low-resolution and lack of model agreement, particularly for precipitation~\cite{schiermeier2010real}.

We can harness information in observational datasets in order to learn statistical models mapping ESM outputs to a higher resolutions. Observational datasets are available through a variety of sources, including satellite observations, station observations, and a mixture of both, namely reanalysis datasets. Often, SD models are built to downscale ESMs directly to a observational station while others aim to downscale to a grid based dataset. Gridded observational datasets are often built by aggregating station observations to a defined grid. For example, in our application, we obtain precipitation through the PRISM dataset at a 4km daily spatial resolution which aggregates station observations to a grid with physical and topographical information~\cite{daly2008physiographically}. We then upscale the precipitation data to $1/8^{\circ}$ (~$12.5$ km) as our high-resolution observations. Following, we upscale further to $1^{\circ}$ corresponding to a low-resolution precipitation, as applied in~\cite{pierce2014statistical}. The goal is then to learn a mapping between our low-resolution and high-resolution datasets.

\begin{figure}
  \centering
  \includegraphics[width=0.5\textwidth]{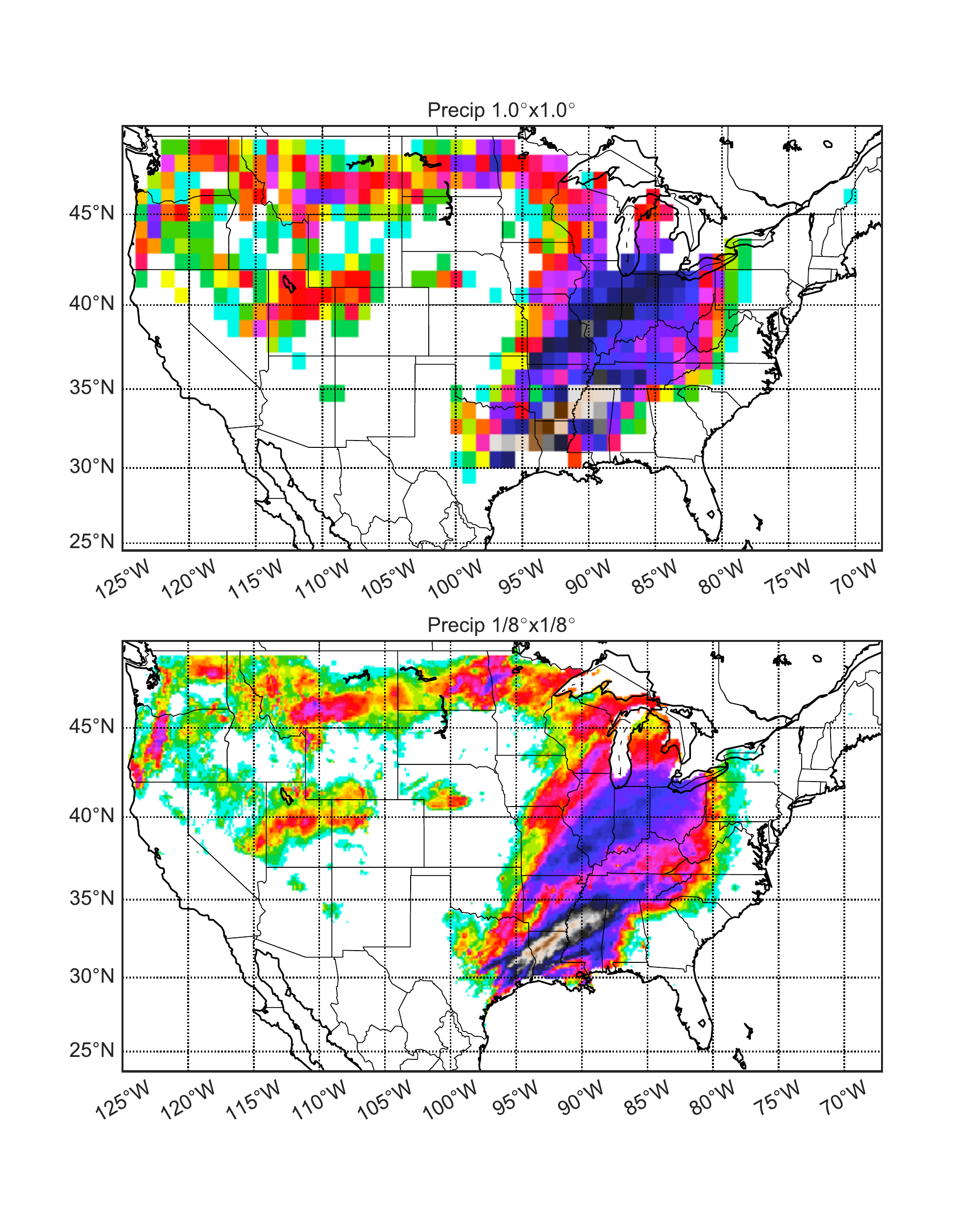}
  \caption{Prism Observed Precipitation: A) Low resolution at $1.0^{\circ}$ ($\sim 100$km). B) High resolution at $1/8^{\circ}$ ($\sim 12.5$km).}
  \vspace{-0.5cm}
\label{fig:lr-hr}
\end{figure}

Furthermore, topography has large effects on weather and climate patterns with lower temperatures, more precipitation, and higher winds~\cite{daly2008physiographically,daly1994statistical}. Taking advantage of the valuable topographical information at different scales, where $1/2^{\circ}$ may capture large scale weather patterns while $1/8^{\circ}$ spatial resolution can capture high-resolution precipitation biases. 

Each of the earth science data products discussed inherently possess rich spatial dependencies, much like images. However, traditionally statistical downscaling methods, particularly regression based models, vectorize spatial data, removing this spatial structure. While colored images contain channels consisting of, for example, red, green, and blue, climate data may be represented analogously such that the channels correspond to climate variables and topographical data. Similar approaches have been applied to satellite datasets for image classification~\cite{basu2015deepsat} and resolution enhancement~\cite{zhang2016deep}. Though climate data is more complex than images due to it's dynamics and chaotic nature, we propose that this representation allows scientists to approach the data in an unconventional manner and apply augmented models developed for image processing. 

\section{Statistical Downscaling}
SD is the problem of mapping a low resolution climate variable to a high resolution projection. This mapping, which must transform a single grid point to multiple points is an ill-posed problem, one with many possible solutions (see Figure~\ref{fig:lr-hr}). However, we can mitigate the ill-posed problem by including static high-resolution topography data in conjunction with other low-resolution climate variables. We learn the SD model using observed climate datasets and then infer downscaled ESM projections. Spatial and temporal non-stationarity of the changing climate system challenges traditional statistical techniques. Downscaling precipitation further challenges these methods with sparse occurrences and skewed distributions. The combination of an ill-posed problem, uncertainty in the climate system, and data sparsity propagates uncertainty in downscaled climate projections further. 

\section{Related Work}
As mentioned previously, SD has a rich and expansive history in the climate community. SD consists of three fundamental categories: regression models and weather classification schemes which improve spatial resolution while weather generators increase temporal resolution (ie. monthly to daily)~\cite{wilby2004guidelines}. As our interest is in increasing spatial resolution we will review regression methods and weather classification. 

Regression methods applied to SD are wide in scope, both linear and non-linear, and vary based on the specific climate variable and temporal scale. For instance, downscaling daily precipitation, which we will focus on, relies on a sparse observational dataset where few days contain rainfall while the amount of rainfall in those days follow a skewed distribution. Automated Statistical Downscaling (ASD) presents a traditional framework for this problem where a classification model is first used to classify days with precipitation followed by a regression to estimate the amount~\cite{hessami2008automated}. Similar approaches, among others, include quantile regression neural networks~\cite{cannon2011quantile}, bayesian model averaging\cite{Zhang2015}, and expanded downscaling\cite{burger1996expanded}. Each of these regression models learns a statistical relationship between observed low- and high-resolution pairs and is then applied to ESMs. Another widely used approached is Bias Corrected Spatial Disaggregation (BCSD), which begins by bias correcting a ESM to match the distribution of the high-resolution observed dataset followed by interpolation and spatial scaling to correct for local biases\cite{wood2004hydrologic,thrasher2012technical}. Though BCSD is a simple approach, it has been shown to perform well compared to more complex methods~\cite{Burger2012,maurer2008utility}. Furthermore, we have shown that BCSD performs similarly, or better, when compared to off-the-shelf ASD approaches~\cite{vandal2017intercomparison}.

Weather classification methods take a different approach to statistical downscaling through nearest neighbor estimates, grouping weather events into similar types. Given a set of observed low- and high-resolution pairs, one can compute a distance measure between an ESM and the low-resolution observations to select the nearest high-resolution estimate. Constructed analogues furthers the method by performing a regression on a group of the nearest neighbor estimates\cite{hidalgo2008}. More advances, but similar approaches, have recently been presented, including Hierarchical Bayesian inference models\cite{manor2015bayesian}. 

While the approaches discussed above are often sufficient in downscaling means, they tend to fail at downscaling extreme events. For instance, ASD approaches perform reasonably well at downscaling average precipitation~\cite{hessami2008automated} but performs poorly at the extremes~\cite{Burger2012}. As discussed by B\"{u}rger et al.~\cite{Burger2012} and Mannshardt-Shamseldin et al.~\cite{mannshardt2010downscaling}, as well as others, specific approaches to downscaling extremes are often required. These specialized approaches, such as those using Generalized Extreme Value theory, have been developed for this purpose~\cite{mannshardt2010downscaling,hashmi2011comparison}. Ideally, a single approach to downscaling leveraging all available information would capture both averages and extremes, giving the user a more credible dataset. 

To our knowledge little work has been attempted to explicitly capture spatial properties for improving downscaled projections. As computer vision approaches are built to exploit the spatial structure of images, we are motivated to understand the applicability of such methods to climate datasets. As introduced previously, we represent climate variables as channels, analogous to images, and model them similarly. However, we note that this presents an analogy and not a direct correspondence. 

Using the analogy between climate datasets and images, we can relate statistical downscaling to image super-resolution, where one aims to learn a mapping from low- to high-resolution image pairs. Specifically, single image super-resolution (SR), as the name suggests, increases the resolution of a single image, rather than multiple images, from a scene. 

The most successful approaches to SR have been shown to be patch based (or example-based) techniques, achieving state-of-the-art performance~\cite{timofte2014a,dong2014learning,wang2015deep}. Originally proposed by Glasner et al.~\cite{glasner2009super}, patch based methods exploit self-similarity between images to produce exemplar patches. This approach has evolved into different variations of nearest neighbor techniques between low- and high-resolution patches through what is known as dictionary learning~\cite{freeman2002example,chang2004super,yang2010image}. Dictionary learning approaches to SR are analogous to those presented by weather classification SD schemes. Furthermore, approaches including kernel regression~\cite{yang2013fast}, random forests\cite{schulter2015fast}, and anchored neighborhood regression~\cite{timofte2014a}, have been proposed for SR to improve accuracy and speed, all related to methods presented in SD literature~\cite{Ghosh2010,burrows1995cart,stoner2013asynchronous}. Sparse-coding techniques, a form of dictionary learning, have recently shown state-of-the-art results in both speed and accuracy~\cite{timofte2014a}. 

Convolutional neural networks were recently presented as a generalization of sparse coding, improving upon past state-of-the-art performances\cite{dong2014learning,wang2015deep}. The sparse coding generalization, non-linearity, network flexibility, and scalability to large datasets presents an opportunity to apply Super Resolution Convolutional Neural Networks to SD\cite{dong2014learning}.

\section{Methodology}

\newcommand{\argmin}{\operatornamewithlimits{argmin}}

This section begins by describing and formulating Super-Resolution Convolutional Neural Networks (SRCNN), as presented by~\cite{dong2014learning}. We then introduce a stacked SRCNN architecture such that the output of one SRCNN is the input to the following SRCNN. DeepSD, the adaptation of a stacked SRCNNs to SD, is then introduced.

\begin{figure*}
  \centering
  \includegraphics[width=0.9\textwidth]{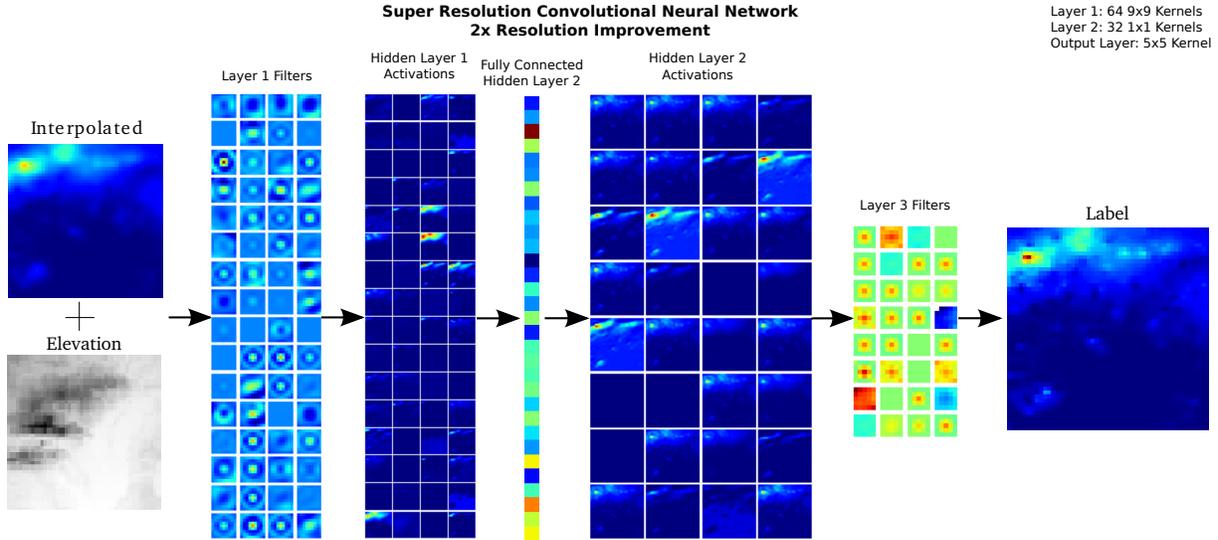}
  \caption{Augmented SRCNN Architecture. From the left to right: Precipitation and Elevation sub-image pair, filters learned in layer 1, layer 1 activations, layer 2 filters, layer 2 activations, layer 3 filters, and HR precipitation label.}
    \vspace{-0.2cm}
\label{fig:filters-activations}
\end{figure*}

\subsection{Super-resolution CNN}

SR methods, given a low-resolution (LR) image, aim to accurately estimate a high-resolution image (HR). As presented by Dong et al.~\cite{dong2014learning}, a CNN architecture can be designed to learn a functional mapping between LR and HR using three operations, patch extraction, non-linear mappings, and reconstruction. The LR input is denoted as $\mathbf{X}$ while the HR label is denoted as $\mathbf{Y}$. 

A three layer CNN is then constructed as follows to produce a high resolution estimate and presented in Figure~\ref{fig:filters-activations}. Layer 1 is formulated as $$F_1(\mathbf{X}) = max(0, W_1 * \mathbf{X} + B_1),$$ where `$*$' is the convolution operation and the $max$ operation applies a Rectified Linear Unit~\cite{nair2010rectified} while $W_1$ and $B_1$ are the filters and biases, respectively. $W_1$ consists of $n_1$ filters of size $c \times f_1 \times f_1$. The filter size, $f_1 \times f_1$, operates as an overlapping patch extraction layer where each patch is represented as a high-dimensional vector. 

Correspondingly, layer 2 is a non-linear operation such that $$F_2(\mathbf{X}) = max(0, W_2 * F_1(\mathbf{X}) + B_2)$$ where $W_2$ consists of $n_2$ filters of size $n_1 \times f_2 \times f_2$ and $B_2$ is a bias vector. This non-linear operation maps high-dimensional patch-wise vectors to another high-dimensional vector. 

A third convolution layer is used to reconstruct a HR estimate such that $$F(\mathbf{X}) = W_3 * F_2(\mathbf{X}) + B_3.$$ Here, $W_3$ contains $1$ filter of size $n_2 \times f_3 \times f_3$. The reconstructed image $F(\mathbf{X})$ is expected to be similar to the HR image, $\mathbf{Y}$
 
This end-to-end mapping then requires us to learn the parameters $\Theta = \{W_1, W_2, W_3, B_1, B_2, B_3 \}$. A Euclidean loss function with inputs $\{\mathbf{X}_i\}$ and labels $\{\mathbf{Y}_i\}$ is used where the optimization objective is defined as: 
\begin{equation} \label{eq:srcnn-loss}
\argmin_{\Theta} \sum_{i=1}^n \Vert F(\mathbf{X}_i;\Theta) - \mathbf{Y}_i \Vert_2^2
\end{equation}
such that $n$ is the number of training samples (batch size).

We note that convolutions in layers 1, 2, and 3 decrease the image size depending on the chosen filter sizes, $f_1$,$f_2$, and $f_3$. At test time, padding using the replication method is applied before the convolution operation to ensure the size of the prediction and ground truth correspond. During training, labels are cropped such that $\mathbf{Y}$ and $F(\mathbf{X}_i ; \Theta)$, without padding, are of equal size.

\subsection{Stacked SRCNN}

Traditional SR methods are built for resolution enhancements of factors from 2 to 4 while statistical downscaling conservatively requires resolution increases of factors from 8 to 12. Rather than enhancing resolution directly to 8-12x, as SR applications typically do, we take an alternative approach. To achieve such a large resolution improvement, we present stacked SRCNNs such that each SRCNN increases the resolution by a factor of $s$. This approach allows the model to learn spatial patterns at multiple scales, requiring less complexity in the spatial representations. The approach of stacking networks has been widely used in deep learning architectures, including stacked denoising autoencoders~\cite{vincent2010stacked} and stacked RBMs for deep belief networks~\cite{hinton2006reducing}. However, contrary to the above networks where stacking is applied in an unsupervised manner, each SRCNN is trained independently using their respective input/output resolutions and stacked at test time. 

A similar approach using cascading super-resolution networks showed positive results for upscaling factors below 4~\cite{wang2015deep}, however through experimentation we found that cascading SRCNNs performed worse than stacked SRCNNs. The ability of arbitrarily upscaling ground truth images to lower resolution allows for input/output pairs to be produced at multiple scales to train stacked SRCNNs. However, while training a cascading model, the output of each SRCNN is the input to the following SRCNN, which may be leading to unnecessary error propagation through the network. 

\subsection{DeepSD}

\begin{figure}
\centering
\includegraphics[width=0.5\textwidth]{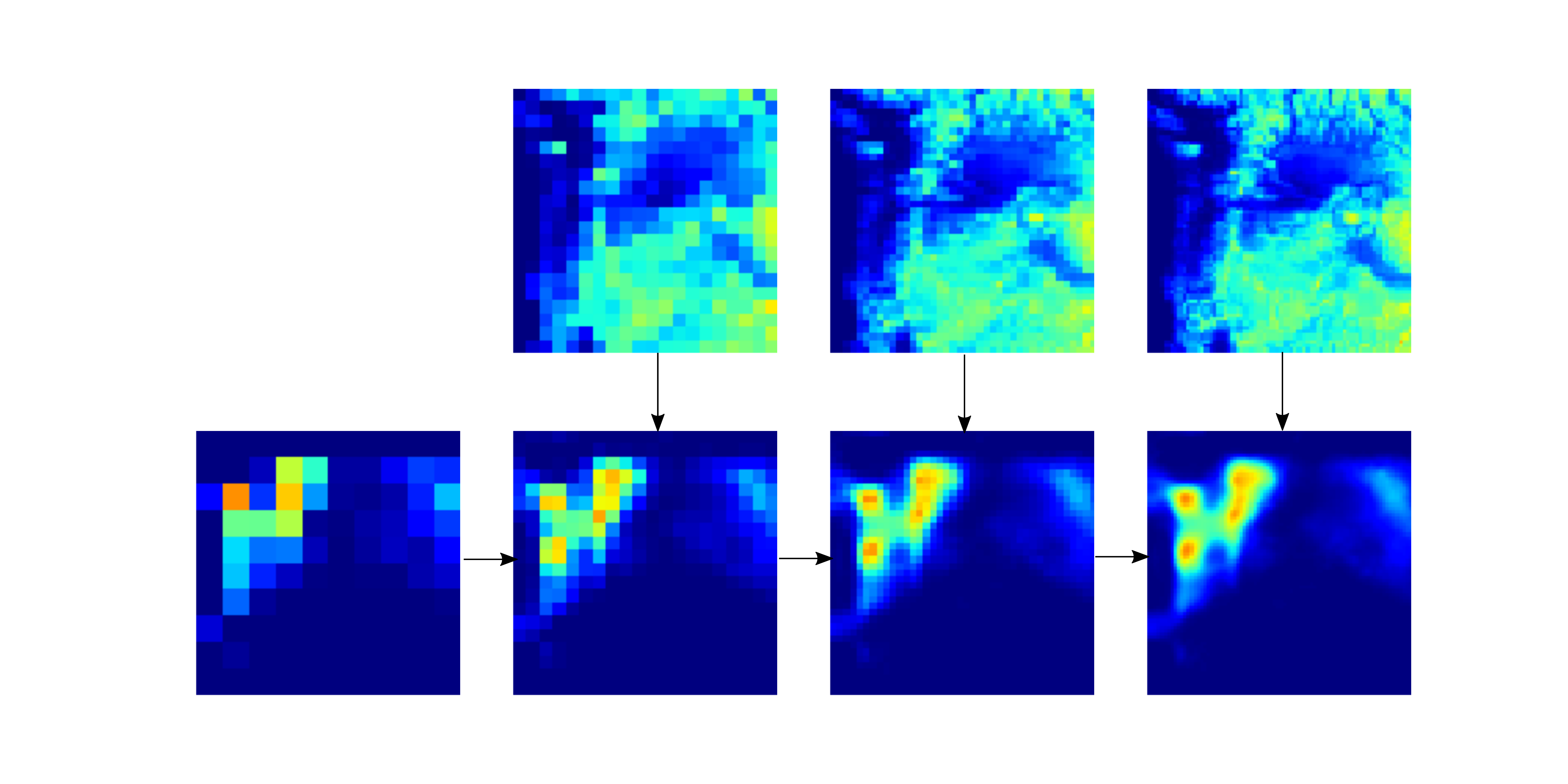}
\caption{Layer by layer resolution enhancement from DeepSD using stacked SRCNNs. Top Row: Elevation, Bottow Row: Precipitation. Columns: 1.0$^{\circ}$, 1/2$^{\circ}$, 1/4$^{\circ}$ and 1/8$^{\circ}$ spatial resolutions.}
  \vspace{-0.5cm}
\label{fig:deepsd}
\end{figure}

We now present DeepSD, an augmented and specific architecture of stacked SRCNNs, as a novel SD technique. When applying SR to images we generally only have a LR image to estimate a HR image. However, during SD, we may have underlying high-resolution data coinciding with this LR image to estimate the HR images. For instance, when downscaling precipitation we have two types on inputs including LR precipitation and static topographical features such as HR elevation and land/water masks to estimate HR precipitation. As topographical features are known beforehand at very high resolutions and generally do not change over the period of interest they can be leveraged at each scaling factor. As done when training stacked SRCNNs, each SRCNN is trained independently with it's associated input/output pairs. As presented in figure~\ref{fig:deepsd}, inference is executed by starting with the lowest resolution image with it's associated HR elevation to predict the first resolution enhancement. The next resolution enhancement is estimated from the previous layer's estimate and it's associated HR elevation. This process is repeated for each trained SRCNN. Figure~\ref{fig:deepsd} illustrates this process with a precipitation event and it's various resolution improvements. We see that this stacked process allows the model to capture both regional and local patterns.

\section{Application of DeepSD}
Though high resolution precipitation is crucial to climate adaptation, it makes up two of the four major holes in climate science~\cite{schiermeier2010real}. Furthermore, both statistical and dynamical downscaling approaches have been shown to add little information beyond coarse ESMs when applied to precipitation~\cite{kerr2013forecasting,Burger2012}. This motivates our application to downscale daily precipitation over the CONUS, a region where data is credible and abundant at high resolutions. As presented above, we use daily precipitation from the PRISM dataset~\cite{daly2008physiographically} and elevation from Global 30 Arc-Second Elevation Data Set (GTOPO30) provided by the USGS. These datasets are used to train and test DeepSD, which we compare to BCSD, a widely used statistical downscaling technique, as well as three off-the-shelf machine learning regression approaches. The years 1980 to 2005 were used for training (9496 days) while the years 2006 and 2014 (3287 days) were used for testing. Lastly, we present a scalable framework on the NASA Earth Exchange (NEX) platform to downscale 20 GCMs for multiple emission scenarios.

\subsection{Training DeepSD}
Our experiments downscale daily precipitation from $1.0^{\circ}$ to $1/8^{\circ}$, an 8x resolution enhancement, using three SRCNN networks each providing a 2x resolution increase ($1.0^{\circ} \rightarrow 1/2^{\circ} \rightarrow 1/4^{\circ} \rightarrow 1/8^{\circ}$).

\noindent
\subsubsection*{Data preprocessing} Data for a single day at the highest resolution, $1/8^{\circ}$, covering CONUS is an ``image'' of size 208x464. Precipitation and elevation are used as input channels while precipitation is the sole output. Images are obtained at each resolution through up-sampling using a bicubic interpolation. For instance up-sampling to $1.0^{\circ}$ decreases the image size from 208x464 to 26x58. Precipitation features for the first SRCNN, downscaling from $1.0^{\circ}$ to $1/2^{\circ}$, are first up-sampled to $1.0^{\circ}$ and then interpolated for a second time to $1/2^{\circ}$ in order to correspond to the output size of 52x116. This process is subsequently applied to each SRCNN depending on it's corresponding resolution. During the training phase, 51x51 sub-images are extracted at a stride of 20 to provide heterogeneity in the training set. The number of sub-images per year (1095, 9125, and 45,625) increase with resolution. Features and labels are normalized to zero mean and unit variance. 

\noindent
\subsubsection*{Training Parameters} All SRCNNs are trained with the same set of parameters, selected using those found to work well by Dong et al.~\cite{dong2014learning}. Layer 1 consists of 64 filters of 9x9 kernels, layer 2 consists of 32 fully connected neurons (1x1 filters), and the output layer uses a 5x5 kernel (see Figure~\ref{fig:filters-activations}). Higher resolution models which have a greater number of sub-images may gain from larger kernel sizes and an increased number of filters. Each network is trained using Adam optimization~\cite{kingma2014adam} with a learning rate of $10^{-4}$ for the first two layers and $10^{-5}$ for the last layers. Each model was trained for $10^{7}$ iterations with a batch size of 200. Tensorflow~\cite{abadi2016tensorflow} was utilized to build and train DeepSD. Training harnessed three Titan X GPUs on an NVIDIA DIGITS DevBox by independently training each SRCNN. Inference was then executed sequentially on a single Titan X GPU on the same machine. 

\subsection{Comparison}
\begin{table}
  \centering
  \begin{tabular}{lrrrrr}
  \toprule
  {} &   Bias &  Corr &  RMSE &  Skill$^{1}$ & Runtime\\
  Model & {\tiny (mm/day)} &   & {\tiny (mm/day)} &      & (secs) \\
  \midrule
  Lasso  &  0.053 & 0.892 & 2.653 &  0.925 & 1297 \\
  ANN    &  0.049 & 0.862 & 3.002 &  0.907 & 2015 \\
  SVM    & -1.489 & 0.886 & 3.205 &  0.342 & 27800\\
  BCSD   & -0.037 & 0.849 & 4.414 &  \textbf{0.955} &  \textbf{13} \\
  DeepSD &  \textbf{0.022} & \textbf{0.914} & \textbf{2.529} &  0.947 & 71 \\
  \bottomrule
  \end{tabular}
  \caption{Comparison of predictive ability between all five methods for 1000 randomly selected locations in CONUS. Runtime is computed as the amount of time to downscale 1 year of CONUS.}
  \vspace{-0.5cm}
  \label{tab:asd-compare}
\end{table}

\footnote[1]{See section 6.2.2 for details.}

\begin{figure*}[t]
\centering
\includegraphics[width=\textwidth]{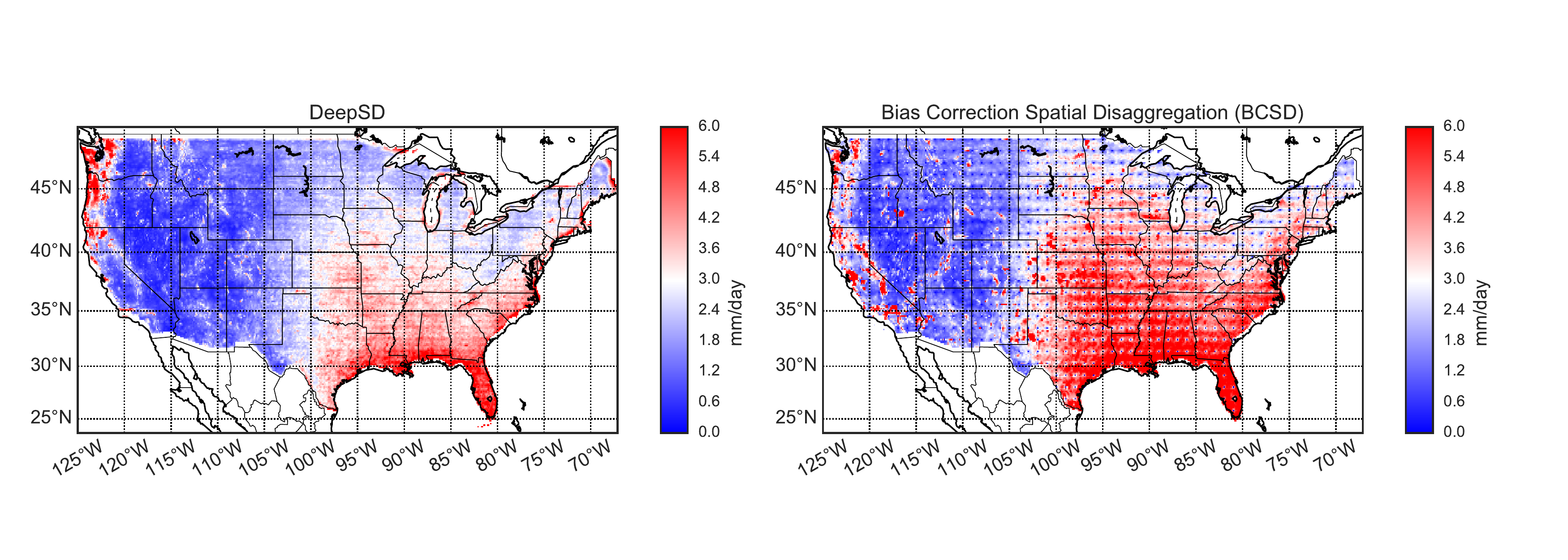}
\caption{Daily Root Mean Square Error (RMSE) computed at each location for years 2006 to 2014 (test set) in CONUS for A) DeepSD and B) BCSD. Red corresponds to high RMSE while blue corresponds to low RMSE.}
  \vspace{-0.2cm}
\label{fig:rmse-map}
\end{figure*}

\subsubsection{State-of-the-Art Methods}
Bias Correction Spatial Disaggregation (BCSD)~\cite{wood2004hydrologic} is a simple but effective method for statistical downscaling. Many studies have compared BCSD to a variety of other downscaling methods and have found good results in estimating the underlying distribution of precipitation~\cite{Burger2012}. In our experiments we apply the daily BCSD technique~\cite{thrasher2012technical} to precipitation over CONUS. First, the high-resolution precipitation is linearly interpolated to the low-resolution grid. Contrary to applying BCSD to ESMs, we are not required to perform quantile mapping as the distributions are identical. Next, the low-resolution precipitation is interpolated back to high-resolution such that the fine-grained information is lost. Then, scaling factors are computed by dividing high-resolution observations with the interpolated data over the training set (1980-2005). Lastly, the interpolated data is multiplied by the scaling factors to provide downscaled projections. For a more detailed description of this implementation of BCSD see~\cite{thrasher2012technical}. The projections over the test set (2006-2014) are then used for comparison to BCSD.

A second set of methods, Automated-Statistical Downscaling (ASD)~\cite{hessami2008automated}, is applied to compare a variety of regression techniques to DeepSD. ASD consists of two steps for downscaling precipitation: 1. Classifying rainy/non-rainy days ($\>$mm), 2. Estimating total precipitation on rainy days. Hence, this approach requires both classification and regression methods. We compare three ASD approaches using logistic and lasso regression, support vector machine (SVM) classifier and regression, and artificial neural network (ANN) classifier and regression. The Lasso penalty parameter at each location was chosen using 3-fold cross-validation. The SVMs were trained with a linear kernel and a penalty parameter of $1.0$. Each ANN consists of a single layer of 100 units connected with a sigmoid function. A 9 by 9 box for the LR precipitation surrounding the downscaled location is selected as features. Each downscaled location requires individually optimized parameters making the process computationally intensive and complex. All features and labels are normalized to zero mean and unit variance. Hence, we randomly selected 1000 locations to downscale as a trade-off between complexity and statistical certainty around our results. 

\subsubsection{Daily Predictability}

\begin{table}
\centering
  \begin{tabular}{llrrrr}
  \toprule
      &      &  Bias &  Corr &  RMSE &  Skill$^{1}$ \\
  Season & Model & {\tiny (mm/day)} &   & {\tiny (mm/day)} &        \\
  \midrule
  DJF & BCSD &  \textbf{0.02} &         0.89 &  2.36 &   0.95 \\
      & DeepSD & -0.03 &         \textbf{0.95} &  \textbf{1.53} &   \textbf{0.94} \\
  JJA & BCSD &  \textbf{0.01} &         0.78 &  4.15 &   \textbf{0.92} \\
      & DeepSD & -0.05 &         \textbf{0.86} &  \textbf{3.29} &   0.91 \\
  MAM & BCSD &  \textbf{0.01} &         0.87 &  3.02 &   \textbf{0.94} \\
      & DeepSD & -0.03 &         \textbf{0.93} &  \textbf{2.29} &   0.93 \\
  SON & BCSD &  \textbf{0.01} &         0.87 &  3.27 &   \textbf{0.94} \\
      & DeepSD & -0.04 &         \textbf{0.93} &  \textbf{2.31} &   \textbf{0.94} \\
  \bottomrule
  \end{tabular}
  \caption{Comparison of Predictive Ability between DeepSD and BCSD for each season, Winter, Summer, Spring, and Fall. Values are computed at each location in CONUS and averaged.}
  \vspace{-0.5cm}
  \label{tab:seasonal}
\end{table}

\begin{figure*}[t]
\centering
\includegraphics[width=\textwidth]{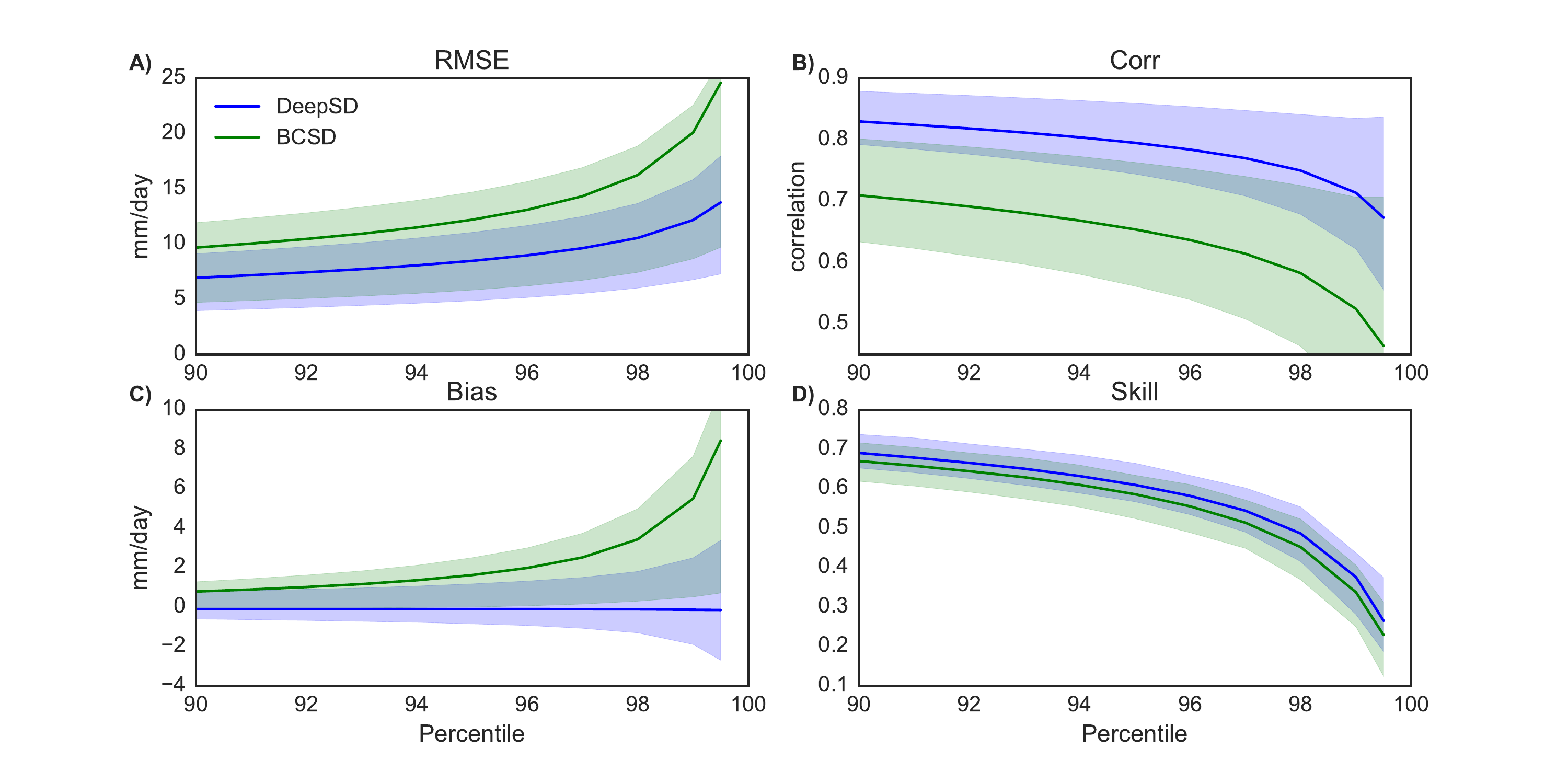}
\caption{Comparison of DeepSD and BCSD for increasingly extreme precipitation. At each location in CONUS, all precipitation events above a percentile threshold (x-axis) are selected. Percentile thresholds between 90 and 99.9 are used. A) RMSE, B) Correlation, C) Bias, and D) Skill are computed at each location and averages over CONUS. The confidence bounds of each metric are taken from the 25th and 75th quantiles.}
\label{fig:extremes}
\end{figure*}

DeepSD's ability to provide credible projections is crucial to all stakeholders. While there are many facets to statistical downscaling, we use a few key metrics to show DeepSD's applicability. Root mean square error (RMSE) and Pearson's correlation are used to capture the predictive capabilities of the methods. Figure~\ref{fig:rmse-map} maps this RMSE (mm/day) for each location. Bias, the average error, presents the ability to estimate the mean while a skill score metric, as presented in~\cite{perkins2007evaluation}, is used to measure distribution similarity. Skill is computed as
\begin{equation}
skill = \sum_{i=1}^n minimum(Z_o^{(i)}, Z_m^{(i)})
\end{equation}
such that $Z_o$ and $Z_m$ are the observed and DeepSD's empirical probability density function while n is the number of bins. Hence, the skill score is between 0 and 1 where 1 is the best. 

Our first experiment compares five approaches, DeepSD, BCSD, Lasso, SVM, and ANN, on their ability to capture daily predictability, presented in Table~\ref{tab:asd-compare}. The four metrics discussed above are computed and averaged over the 1000 randomly selected locations in CONUS where ASD methods were trained. We find that DeepSD outperforms the other approaches in terms of bias, correlation, and RMSE and closely behind BCSD in terms of skill. Furthermore, we find that SVM performs poorly in testing while having the longest runtime. Similarly, the least complex ASD method, Lasso, outperforms the more complex ANN. As expected, BCSD, a method built around estimating the underlying distribution, does well in minimizing bias and estimating the underlying distribution. For these reasons, in conjunction with our previous findings~\cite{vandal2017intercomparison}, the remaining experiments will limit the methods to DeepSD and BCSD.

In the next experiment compare DeepSD and BCSD, the two scalable and top performing methods from the previous experiment, with each metric over CONUS. Each metric is computed per location and season using the daily observations and downscaled estimates then averaged over CONUS, as presented in Table
~\ref{tab:seasonal}. We find that DeepSD has high predictive capabilities for all seasons, higher correlation and lower RMSE, when compared to BCSD. Similar results are shown in Figure~\ref{fig:rmse-map} where DeepSD has a lower RMSE than BCSD for 79\% of CONUS. Furthermore, we find each method's ability to estimate the underlying distribution well with low bias, $<0.5\text{ mm/day}$, and a high skill score of $\sim0.98$. As BCSD is built specifically to minimize bias and match the underlying distribution, DeepSD's performance is strong. Overall, DeepSD outperforms BCSD for the chosen set metrics.

\subsubsection{Predicting Extremes}

As discussed in section 4, downscaling both averages and extreme events with a single method is challenging. Our last experiment tests this challenge by comparing DeepSD's ability to estimate extreme precipitation events when compared to BCSD, an approach shown to perform well~\cite{Burger2012}. A varying quantile threshold approach is used to test each methods ability to capture extreme events. For instance, given a downscaled location we compute RMSE, correlation, bias, and skill for all precipitation events greater the 90th percentile. This is done for a range of percentiles between 90 and 99.9 and averaged over all locations in CONUS. Along with the mean, we select the 25th and 75th quantiles of each metric over CONUS and plot them as confidence bounds in Figure~\ref{fig:extremes}. Figure~\ref{fig:extremes} presents BCSD's loss of predictive capability when compared to DeepSD. We find that BCSD over-estimates extremes at upper quantiles while DeepSD is relatively stable. Though RMSE, Corr, and Skill becomes worse at these extremes, DeepSD consistently outperforms BCSD, most often with thinner confidence bounds. These results show DeepSD's ability to perform well for increasingly extreme precipitation events. DeepSD's performance is impressive given that literature has shown that traditional techniques tend to fail when downscaling averages and extremes simultaneously. We hypothesize that capturing nearby spatial information allows DeepSD to isolate areas where extreme precipitation events are more likely than others.

\subsection{Scalability on NASA's NEX}

Comprehensive studies of climate change requires much more than a single ESM simulation but rather multiple projections from different models, emission scenarios, and initial conditions in order to capture uncertainty. In total, CMIP5 contains more than 20 models at 4 emission scenarios (Representative Concentration Pathways (RCPs) 2.6,4.5,6.5,8.6), a variable number of initial conditions, and multiple climate variables at a daily temporal resolution. Generally, each prospective projection is available from 2006 to 2100 while retrospective projections are available from 1850 to 2005. Limiting the downscaled projections to encompass CONUS at $1/8^{\circ} \times 1/8^{\circ}$, a single simulation requires 134MB. Following the current timeframes of downscaled projections on NASA's NEX platform, downscaling from the year 1950 to 2005 requires 7.4GB of storage while each prospective run needs 13GB. Hence, the final dataset size is 1.2TB. When downscaled further to PRISM's native resolution, $1/16^{\circ} \times 1/16^{\circ}$, the dataset size increases to approximately 5TB. Furthermore, the dataset scales linearly as more variables are added, including temperature minimum and maximum.

We test computational scalability by computing the amount of time taken to downscale 1 year of CONUS, presented in Table~\ref{tab:asd-compare}. For DeepSD, this includes the 3 feed-forward processes and their corresponding interpolations computed on a single GPU in NVIDIA DIGITS DevBox. Runtime for each of the ASD methods is estimated and scaled from the length of time to downscale the 1000 selected locations using 40 CPUs in parallel (Intel Xeon CPU E5-2680 2.8GHz). BCSD's runtime is computed as the amount of time taken to quantile map, interpolate, and scale a years worth of data. We find that BCSD and DeepSD widely outperform the ASD approaches. While BCSD provides the quickest runtime, DeepSD is still scalable. 

Though DeepSD is a highly scalable method, due to a single feed-forward neural network architecture, generating such large datasets still requires heavy computational power. However, storage and compute resources are satisfied by dedicated access to the Pleiades supercomputer housed in NASA's Advanced Supercomputer Division (HECC) at NASA Ames. High resolution projections can be quickly computed using GPU's, which are available on each node, in coordination with the Message Passing Interface (MPI). High-resolution projections are then stored on NEX's filesystem which has currently 2.3PB of rapid-access  storage in addition to large scale tape storage accessible on the HECC platform. In this paper we present a methodology for statistical downscaling, DeepSD, that leverages recent advances in image super-resolution and convolutional neural networks. DeepSD differs from previous SD methods by explicitly capturing spatial structure while improving scalability. A brief comparison with baseline SD techniques, BCSD and ASD, shows promising results in predictive capabilities when downscaling precipitation over the continental United States. Lastly, we describe how DeepSD can be scaled using NASA's Earth Exchange platform to provide an ensemble of downscaled climate projections from more than 20 ESMs. 

\section{Conclusion}
Though DeepSD shows promise for SD, there are still some limitations in our experimentation regarding spatial and temporal generalization. An advantage of DeepSD is that a single trained model is able to downscale spatial heterogeneous regions. However, we do not test predictability in regions where the model was not trained. Future work will examine this hypothesis to understand DeepSD's credibility in regions with few observations. Second, we do not test temporal non-stationarity, a longstanding problem in statistical downscaling. Evaluation under non-stationarity can be tested using approaches presented by Salvi et al.~\cite{Salvi2015}, such that training and testing data is split between cold/warm years. As there is a single model for all locations, including cold and warm climates, we hypothesize that DeepSD is capable of capturing non-stationarity.

Furthermore, future work can improve multiple facets of DeepSD. For instance, the inclusion of more variables such as temperature, wind, and humidity at different pressure levels of the atmosphere may capture more climate patterns. Also, downscaling multiple climate variables simultaneously could be explored to find similar spatial patterns in the high-resolution datasets, such as high temperatures and increased precipitation. Most importantly, DeepSD fails to capture uncertainty around its downscaled projections, a key factor in adapting to climate change. Recent advances in Bayesian Deep Learning concepts~\cite{gal2016uncertainty} may aid in quantifying uncertainty. Though these limitations exist, DeepSD is a scalable architecture with high predictive capabilities which provides a novel framework for statistical downscaling.

\begin{acks}
This work was supported by NASA Earth Exchange (NEX), \grantsponsor{}{National Science Foundation CISE Expeditions in Computing}{https://www.nsf.gov/awardsearch/showAward?AWD_ID=1029711} under grant number:~\grantnum{}{1029711}, \grantsponsor{}{National Science Foundation CyberSEES}{https://www.nsf.gov/awardsearch/showAward?AWD_ID=1442728} under grand number:~\grantnum{}{1442728}, and \grantsponsor{}{National Science Foundation BIGDATA}{https://www.nsf.gov/awardsearch/showAward?AWD_ID=1447587} under grant number:~\grantnum{}{1447587}. The GTOPO30 dataset was distributed by the Land Processes Distributed Active Archive Center (LP DAAC), located at USGS/EROS, Sioux Falls, SD. \url{http://lpdaac.usgs.gov}. We thank Arindam Banerjee for valuable comments.
\end{acks}

\bibliographystyle{ACM-Reference-Format}
\bibliography{deepsd}

\end{document}